\documentclass[article]{agujournal}

\journalname{Space Weather}

\usepackage{bm}
\usepackage{amsmath}
\usepackage{subfigure}
\usepackage[colorlinks=true, pdfstartview=FitV, linkcolor=black, citecolor= black, urlcolor= black]{hyperref}
\usepackage{overcite}
\usepackage{footnpag}			      	
\usepackage{algorithm,algorithmic}
\usepackage{multirow,bm}
\usepackage{xcolor}
\usepackage[normalem]{ulem}
 \usepackage{subfigmat}

\bibpunct{(}{)}{;}{a}{}{,}

\begin{document}

\title{Machine-Learned HASDM Model with Uncertainty Quantification}

%
%

 \authors{Richard J. Licata\affil{1}\thanks{1306 Evansdale Drive, Morgantown, West Virginia 26506-6106.}, Piyush M. Mehta\affil{1}, W. Kent Tobiska, and S. Huzurbazar\affil{3}}

\affiliation{1}{Department of Mechanical and Aerospace Engineering,
West Virginia University, Morgantown, West Virginia, USA.}
\affiliation{2}{Space Environment Technologies, Pacific Palisades, California, USA.}
\affiliation{3}{School of Mathematical and Data Sciences,
West Virginia University, Morgantown, West Virginia, USA.}


\correspondingauthor{Richard J. Licata}{rjlicata@mix.wvu.edu}


\begin{keypoints}
\item First thermosphere model with robust and reliable uncertainty estimates is developed.
\item HASDM-ML provides predictions with approximately 10\% error relative to HASDM database across 20 years of available data.
\item Heteroskedastic loss function provides model with well-calibrated uncertainty estimates across all conditions.

\end{keypoints}


\begin{abstract}


The first thermospheric neutral mass density model with robust and reliable uncertainty estimates is developed based on the SET HASDM density database. This database, created by Space Environment Technologies (SET), contains 20 years of outputs from the U.S. Space Force's High Accuracy Satellite Drag Model (HASDM), which represents the state-of-the-art for density and drag modeling. We utilize principal component analysis (PCA) for dimensionality reduction, creating the coefficients upon which nonlinear machine-learned (ML) regression models are trained. These models use three unique loss functions: mean square error (MSE), negative logarithm of predictive density (NLPD), and continuous ranked probability score (CRPS). Three input sets are also tested, showing improved performance when introducing time histories for geomagnetic indices. These models leverage Monte Carlo (MC) dropout to provide uncertainty estimates, and the use of the NLPD loss function results in well-calibrated uncertainty estimates without sacrificing model accuracy (<10\% mean absolute error). By comparing the best HASDM-ML model to the HASDM database along satellite orbits, we found that the model provides robust and reliable uncertainties in the density space over all space weather conditions. A storm-time comparison shows that HASDM-ML also supplies meaningful uncertainty measurements during extreme events.

\end{abstract}

\section*{Plain Language Summary}

The first upper-atmospheric density model with robust and reliable uncertainty estimates is developed based on the SET HASDM density database. This database contains 20 years of outputs from the High Accuracy Satellite Drag Model (HASDM), which represents the state-of-the-art for density and drag modeling. We use a decomposition tool called principal component analysis (PCA) to reduce the dimensionality of the dataset. Three loss functions, mean square error (MSE), negative logarithm of predictive density (NLPD), and continuous ranked probability score (CRPS), are tested with three input sets to identify the best-performing model. We optimize nine models (all three loss functions and input sets) and compare the prediction accuracy  and the reliability of its uncertainty estimates. The models leverage Monte Carlo dropout to generate probabilistic outputs from which we extract model uncertainty. We find that using an input set containing a time series for the geomagnetic indices results in the most accurate models. In addition, the model using these inputs with the NLPD loss function has sufficient performance ($\sim$10\% absolute error) and the most calibrated/reliable uncertainty estimates on independent data. We test this model's uncertainty capabilities in the density space along  satellite orbits from 2002-2010, showing the model's reliability across all conditions.

\section{Introduction}\label{sec:intro}

Space Situational Awareness (SSA) is a major focus of space agencies and private companies worldwide. With the number of objects in low-Earth orbit (LEO) continuously growing, knowledge of future satellite/debris positions is becoming increasingly important \citep{constellations}. While there are numerous perturbations affecting the trajectories of objects, atmospheric drag is the largest source of uncertainty in the LEO regime. Our current understanding of the thermosphere is incomplete, resulting in imperfect modeling of neutral mass density. Over the past several decades, researchers have developed increasingly accurate models and made improvements to existing ones. This has come from a combination of the  incorporation of new measurements and refinements of the underlying physics \citep{Emmert07}.

The thermosphere is the neutral region of the upper atmosphere ranging from $\sim$90 km to 500-1000 km depending on space weather conditions. The thermosphere is impacted by external forces (e.g. space weather events) and internal dynamics (e.g. species transport). Extreme Ultraviolet (EUV) emissions from the Sun heat up the thermosphere forcing expansion and increasing density at a given location. Effects from solar emissions can be represented by different solar indices and proxies (e.g. \textit{F\textsubscript{10.7}}) which serve as reliable model drivers \citep{dev_inds}. The Sun continuously emits particles in the form of solar wind which interact with and disrupt the Earth's magnetosphere. During eruptive events, such as coronal mass ejections, there are strong interactions between the particles and the magnetosphere which result in energy being deposited into the thermosphere through Joule heating and particle precipitation at high latitudes \citep{Knipp04}. This energy enhancement causes large increases in density and is difficult to model \citep{storms_CME,therm_storm}. The planetary amplitude index (\textit{a\textsubscript{p}}) has 28 discrete values that represent the level of global geomagnetic activity. During geomagnetic storms, the storm-time disturbance index (\textit{Dst}) can be used to improve density modeling \citep{JB2008}. While space weather indices and proxies are useful as model drivers, their relationship with density is imperfect, as is our ability to forecast these drivers \citep{Benchmarking}.

These modeling limitations put a stress on Space Domain Awareness (SDA) which shifts the focus from catalog maintenance to threat assessment. Uncertainties in density modeling can result in large discrepancies between expected and observed satellite positions \citep{Bussy-Virat,DUQ,MUQ}. A major improvement in density modeling capabilities came with the introduction of real-time data assimilation. HASDM \citep{HASDM} is an assimilative model that leverages an empirical model (JB2008) and Dynamic Calibration of the Atmosphere (DCA) to correct the density nowcast with satellite observations. The updated nowcast can then be propagated forward in time. This model/technique represents the current state-of-the-art in our ability to accurately predict atmospheric density. HASDM outputs had not been publicly available until the recent release of the SET HASDM density database \citep{HASDM_SET_Data}. 

The continued commercial satellite launches are creating a necessity for managing the increasing traffic in orbit. A model that balances prediction accuracy and robustness of uncertainty estimates will be a critical tool for the rising importance of Space Traffic Management (STM) \citep{STM}. In this work, we investigate different modeling techniques using machine learning (ML) to create a surrogate for the database. This allows us to extrapolate beyond the limits of the existing database. A straightforward way to accomplish this is by creating a nonlinear regression model on the full density grid. However, this creates limitations in the context of uncertainty quantification (UQ). To combat this, we focus on the development of nonlinear reduced order models (ROMs). For dimensionality reduction, a linear technique called principal component analysis (PCA) is utilized. The predictive models are standard feed-forward neural networks, or artificial neural networks (ANNs), that employ nonlinear activation functions. We explore various custom loss functions in training in order to obtain reliable or calibrated uncertainty estimates, crucial for collision avoidance operations. 

The paper is organized as follows. We first explain the data and the various techniques used for model development. Then, we show the performance and UQ capabilities using a standard mean square error (MSE) loss function. We improve the UQ capabilities with the introduction of the negative logarithm of predictive density (NLPD) and the continuous ranked probability score (CRPS) loss functions. The best HASDM-ML model, in terms of accuracy and reliability, is then compared to the HASDM database in the density space.

\section{Methodology}

\subsection{Data}

HASDM is an assimilative framework using the Jacchia-Bowman 2008 Empirical Thermospheric Density Model (JB2008) as a background density model. HASDM improves upon the density correction work of Marcos et al. \citeyearpar{Marcos} and Nazerenko et al. \citeyearpar{Naz} to modify 13 global temperature correction coefficients with its DCA algorithm. HASDM uses observations of more than 70 carefully chosen calibration satellites to estimate local density values. The satellite orbits span an altitude range of 190-900 km although a majority are between 300 and 600 km \citep{HASDMrev}. The HASDM algorithm uses a prediction filter that employs wavelet and Fourier analysis for the correction coefficients \citep{HASDM}. Another highlight of HASDM's novel framework is its segmented solution for the ballistic coefficient (SSB). This allows the ballistic coefficient estimate to deviate over the fitting period for the satellite trajectory estimation.

SET validates the HASDM output each week and archives the results. The archived values from 2000-2020 make up the SET HASDM density database, upon which this work is based. The database contains density predictions with 15$^\circ$ longitude, 10$^\circ$ latitude, and 25 km altitude increments spanning from 175 - 825 km. This results in 12,312 grid points for every three hours from the start of 2000 to the end of 2019. For further details on HASDM, the reader is referred to Storz et al. \citeyearpar{HASDM}, and for details on SET's validation process and on the database, the reader is referred to Tobiska et al. \citeyearpar{HASDM_SET_Data}. 

JB2008, and subsequently HASDM, use solar and geomagnetic indices/proxies to drive them. \textit{F\textsubscript{10.7}} (referred to as \textit{F\textsubscript{10}} in this work) is a legacy solar proxy used widely in thermospheric density models. It represents 10.7 cm solar radio flux which does not directly interact with the thermosphere, but it is a reliable proxy for solar EUV heating. The \textit{S\textsubscript{10}} index is a measure of the integrated 26-34 nm solar EUV emission which is absorbed by atomic oxygen in the middle thermosphere. \textit{M\textsubscript{10}} is a proxy that represents the Mg II core-to-wing ratio and is a surrogate for far ultraviolet photospheric 160-nm Schumann-Runge Continuum emissions that cause molecular oxygen dissociation in the lower thermosphere. The last solar index used by JB2008 is \textit{Y\textsubscript{10}} which is a hybrid measure of solar coronal X-ray emissions and Lyman-alpha. \textit{S\textsubscript{10}}, \textit{M\textsubscript{10}}, and \textit{Y\textsubscript{10}} have no relation to the 10.7 cm wavelength but are converted to solar flux units (sfu) through linear regression with \textit{F\textsubscript{10}}. The 81-day solar centered averages of these four solar drivers are used as inputs to JB2008 and are denoted by the "\textit{81c}" subscript.

\begin{figure}[htb!]
	\centering
	\small
	\includegraphics[width=0.85\textwidth]{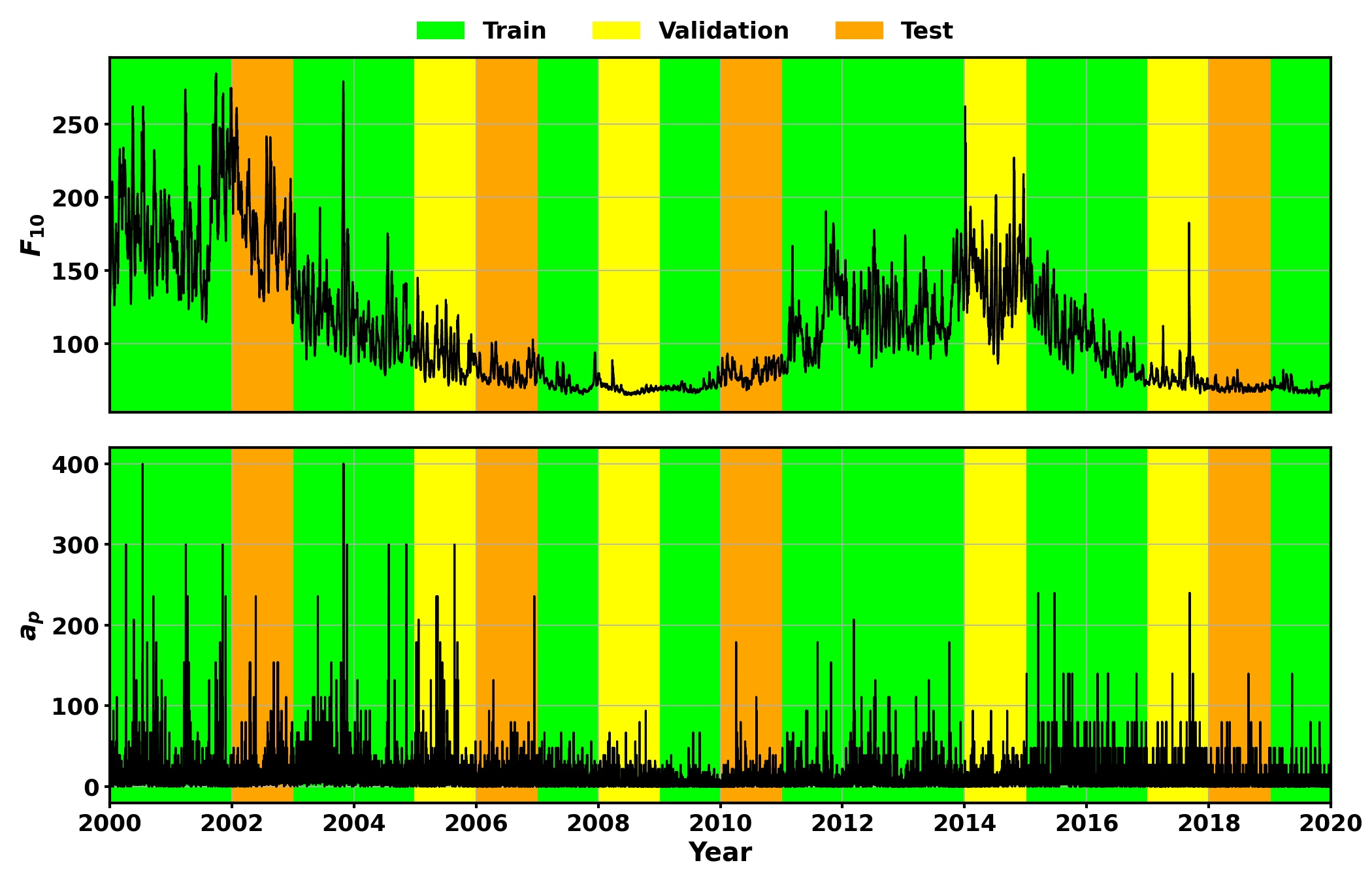}
	\caption{\textit{F\textsubscript{10}} and \textit{a\textsubscript{p}} at available data points shaded to show the training, validation, and test splits.}
	\label{f:TVT}
\end{figure}

For temperature equations corresponding to geomagnetic activity, JB2008 utilizes \textit{a\textsubscript{p}} and \textit{Dst}. The 3-hour \textit{a\textsubscript{p}} is indicative of overall geomagnetic activity and is only measured at low to mid-latitudes. It has 28 discrete values which are capped at 400 \textit{2nT} leading to further limitations. To combat this, JB2008 uses \textit{Dst} during geomagnetic storms if the minimum \textit{Dst} < -75 \textit{nT}, and it is the primary parameter for energy being deposited into the thermosphere during storms. The data (drivers and density) is split into training, validation, and test data using 60\%, 20\%, and 20\%, respectively as shown in Figure \ref{f:TVT}. Table \ref{t:HASDM_samples} shows the number of time steps in the SET HASDM density database across various space weather conditions. The cutoff values for \textit{F\textsubscript{10}} and \textit{a\textsubscript{p}} are obtained from Licata et al. \citeyearpar{Benchmarking} where they were used to bin space weather driver forecasts.

\begin{table}[htb!]
	\fontsize{10}{10}\selectfont
    \caption{Number of time steps for different space weather conditions across the SET HASDM density database.}
   \label{t:HASDM_samples}
        \centering 
   \begin{tabular}{c | c | c | c | c | c} 
      \hline 
          & $\mathbf{F_{10} \leq 75}$ & $\mathbf{75 < F_{10} \leq 150}$ & $\mathbf{150 < F_{10} \leq 190}$ & $\mathbf{F_{10} > 190}$ & \textbf{All} $\mathbf{F_{10}}$ \\ \hline
         $\mathbf{a_p \leq 10}$ & 13,839 & 22,034 & 4,126 & 2,088 & 42,087 \\ \hline
         $\mathbf{10 < a_p \leq 50}$ & 3,003 & 9,226 & 1,982 & 1,091 & 15,302 \\ \hline
         $\mathbf{a_p > 50}$ & 54 & 652 & 196 & 149 & 1,051 \\ \hline
         \textbf{All} $\mathbf{a_p}$ & 16,896 & 31,912 & 6,304 & 3,328 & 58,440 \\ \hline
   \end{tabular}
\end{table}

In Table \ref{t:HASDM_samples}, there is clear under-representation of geomagnetic storms in this vast dataset. This can cause limitations in model development, because over 98\% of the dataset corresponds to \textit{a\textsubscript{p}} $\leq$ 50. Hierarchical modeling could be used for data of this nature, but we proceed with the development of a single comprehensive model.

\subsection{Principal Component Analysis}

Principal Component Analysis is an eigendecomposition technique that determines uncorrelated linear combinations of the data that maximize variance \citep{Pearson_PCA,PCA2}. PCA has been previously used to analyze atmospheric density models and accelerometer-derived density sets. Researchers applied PCA to CHAllenging Minisatellite Payload (CHAMP) and Gravity Recovery and Climate Experiment (GRACE) accelerometer-derived density data in order to analyze the dominant modes of variation and identify phenomona encountered by the satellites \citep{EOF1,EOF2,EOF3}. PCA has also been leveraged for dimensionality reduction to create linear dynamic reduced order models \citep{MehtaROMCal,MehtaROM,RealtimeROM,TLE_ROM}. Licata et al. \citeyearpar{QualHASDM} recently applied PCA to the SET HASDM density database for investigation.

\begin{figure}[htb!]
	\centering
	\small
	\includegraphics[width=0.92\textwidth]{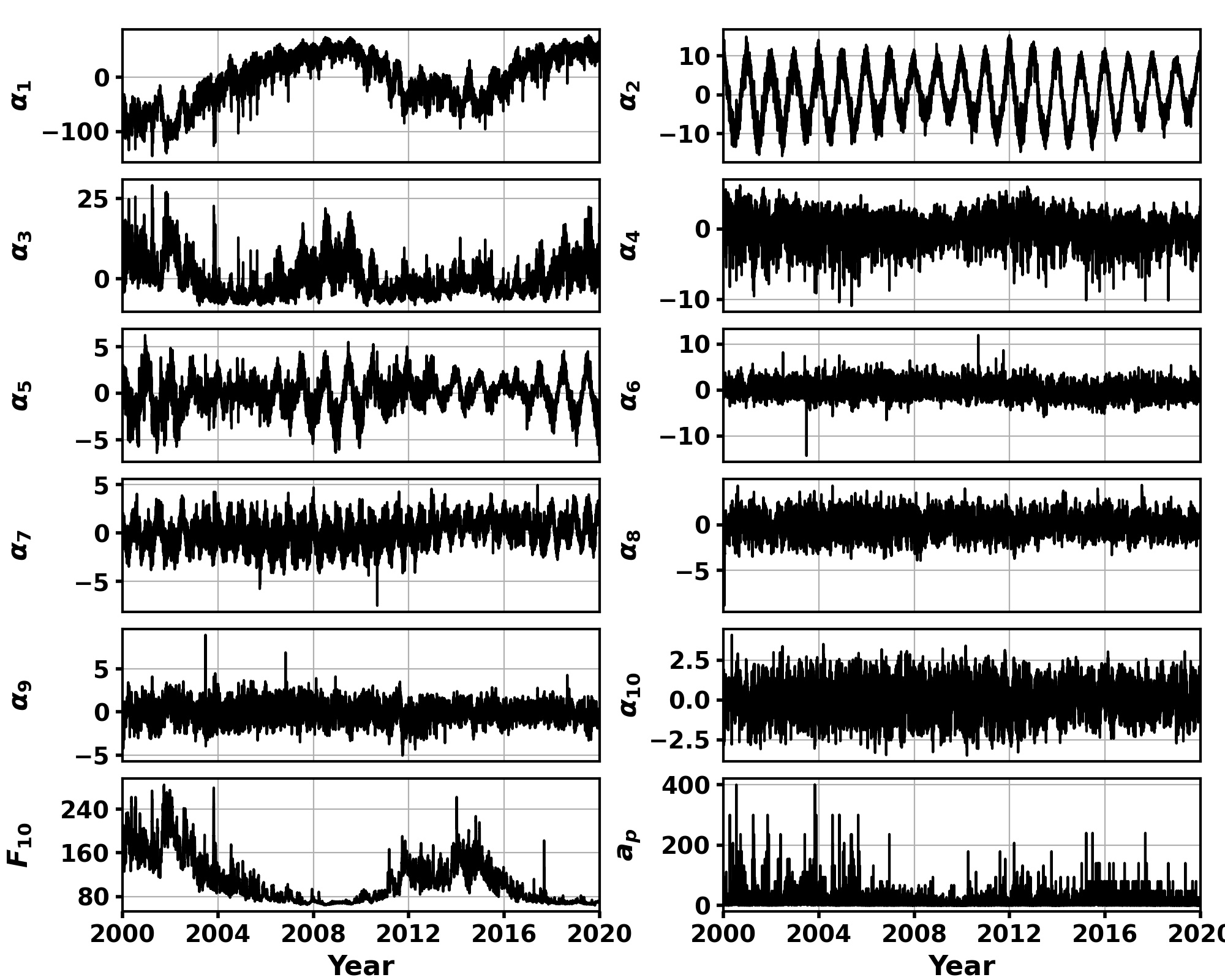}
	\caption{First 10 PCA coefficients ($\alpha_i$) for the HASDM database along with \textit{F\textsubscript{10}} and \textit{a\textsubscript{p}}.}
	\label{f:PCA}
\end{figure}

The SET HASDM dataset is a prime candidate for PCA with the goal of predictive modeling due to its high dimensionality. The three spatial dimensions are first flattened to make the dataset two-dimensional (time $\times$ space). Then a common logarithm (\textit{log\textsubscript{10}}) of the density values is taken in order to reduce the variance of the dataset. Next, we subtract the temporal mean for each cell to center the data and save it for later use. Finally, we perform PCA using the \textit{svds} function in \textit{MATLAB} to acquire the $U$, $\Sigma$, and $V$ matrices (shown in Equation \ref{eq2}). PCA decomposes the data and separates spatial and temporal variations as in Equation \ref{eq1}, 
\begin{equation} \label{eq1}
\begin{split}
\mathbf{x}\left(\mathbf{s},t\right)=\mathbf{\bar{x}}\left(\mathbf{s}\right)+\mathbf{\widetilde{x}}\left(\mathbf{s},t\right)
\;\;\;\textrm{and}\;\;\;
\mathbf{\widetilde{x}}\left(\mathbf{s},t\right)=\sum^r_{i=1}\alpha_i\left(t\right)U_i\left(s\right)
\end{split}
\end{equation}
where $\mathbf{x}\in\mathbb{R}^n$ is the log-transformed model output state (HASDM density on its full 3D grid), $\mathbf{\bar{x}}$ is the mean, $\mathbf{\widetilde{x}}$ is the deviation about the mean, $\mathbf{s}$ represents the spatial dimension, $r$ is the choice of order truncation (here $r$=10), $\alpha_i(t)$ are temporal coefficients, and $U_i$ are orthogonal modes, or basis functions. Equation \ref{eq2} shows how to derive these modes.
\begin{equation} \label{eq2}
\begin{split}
\mathbf{X}=\left[
  \begin{array}{ccccc}
    \vrule & \vrule & \vrule &        & \vrule \\
    \mathbf{\widetilde{x}_1}    & \mathbf{\widetilde{x}_2} & \mathbf{\widetilde{x}_3}  & \ldots & \mathbf{\widetilde{x}_m}    \\
    \vrule & \vrule & \vrule &        & \vrule 
  \end{array}
\right]
\;\;\;\textrm{and}\;\;\;
\mathbf{X} = U \Sigma V^T
\end{split}
\end{equation}
In Equation \ref{eq2}, $m$ represents the ensemble size (two solar cycles with HASDM) and $\mathbf{X}$ represents the log-transformed, mean-centered data . The left unitary matrix, $U$, is made of orthogonal vectors that represent the  modes of variation, and $\Sigma$ is a diagonal matrix consisting of the squares of the eigenvalues that correspond to the vectors in $U$. We encode the data ($\mathbf{X}$) into temporal coefficients ($\alpha_i$) by performing matrix multiplication between $\Sigma$ and $V^T$. To decode back to the density, the coefficients are multiplied by $U$ with the temporal mean added at each cell, and taking the antilogarithm ($10^y$) of the resulting value completes the back-transformation. Figure \ref{f:PCA} shows the first 10 PCA coefficients generated using the methods described above. The first five are discussed in detail by Licata et al. \citeyearpar{QualHASDM}.

\subsection{Hyperparameter Tuning} \label{sec:tuner}

In machine learning, developing an accurate model previously required a manual architecture selection process that did not guarantee optimal performance for a given application. Recent developments in hyperparameter tuning make this process much quicker and more thorough. In this work, we use KerasTuner which allows the user to define the ranges and choices of any hyperparameter and to choose a search algorithm \citep{KT}. We train all models with 100 trials (or architectures) with the first 25 being randomly sampled from our hyperparameter space while the last 75 trials take advantage of the tuner's Bayesian optimization scheme.

The tuner trained three identical models for each trial and returned the lowest validation loss after 100 training iterations, or epochs. The Bayesian optimization scheme chooses future architectures based on the validation losses resulting from previous trials. Once completed, the tuner returns the best 10 models which we can evaluate on all three subsets of our data to find the most accurate and generalized model.

\subsection{Regression Modeling}\label{sec:regression}

A traditional approach to regression modeling with UQ is to develop Gaussian process (GP) models \citep{GPR,GPR_Dst}. In the early stages of model development, we attempted this approach -- training GP regression models on each of the PCA coefficients. However, we could only use 10 years of data for training, limited by computational resources, which resulted in higher predictive error. In addition, the ten resulting models were 6.83 GB each making subsequent work cumbersome. Therefore, the results only pertain to the feedforward neural networks with MC dropout. There are three methods we explore in developing the best HASDM-ML model. First, we create a ROM using PCA for dimensionality reduction and a nonlinear ANN for prediction. We then modify this technique using a custom loss function in an attempt to obtain a model with calibrated uncertainty estimates. We test two loss functions to achieve this: NLPD and CRPS; details of which are given in the following section.

ANNs have two key components: features (or inputs) and labels (or outputs). We try using three separate input sets for our regression models, the first two are explained in Table \ref{t:inputs}. The first set is the JB2008 input set, referred to as JB. Licata et al. \citeyearpar{QualHASDM} showed that the SET HASDM density database contained evidence of post-storm cooling mechanisms which cannot be captured solely by geomagnetic indices at epoch. Therefore, we introduce a second set that is similar to the first but with a time history of the geomagnetic indices. We refer to this as JB\textsubscript{H} (Historical JB2008). Unlike the actual JB2008 inputs, all of our input sets contain sinusoidal transformations to the day of year (doy) and universal time (UT) inputs. The four resulting temporal inputs (shown in Equation \ref{eq3}) represent annual (\textit{t\textsubscript{1}} and \textit{t\textsubscript{2}}) and diurnal (\textit{t\textsubscript{3}} and \textit{t\textsubscript{4}}) variations. The use of cyclical functions of time as inputs has been previously demonstrated \citep{EXTEMPLAR,EXTEMPLAR-ML}.

\begin{equation} \label{eq3}
\begin{split}
t_1=sin\left(\frac{2\pi doy}{365.25}\right),
\;\;\;\;
t_2=cos\left(\frac{2\pi doy}{365.25}\right),
\;\;\;\;
t_3=sin\left(\frac{2\pi UT}{24}\right),
\;\;\;\;
t_4=cos\left(\frac{2\pi UT}{24}\right).
\end{split}
\end{equation}

\begin{table}[htb!]
	\fontsize{10}{10}\selectfont
    \caption{List of inputs in the two sets used for model development.}
   \label{t:inputs}
        \centering 
   \begin{tabular}{c | c | c | c | c | c } 
      \hline 
          \multicolumn{3}{c |}{\textbf{JB2008 Inputs}} & \multicolumn{3}{c }{\textbf{Historical JB2008 Inputs}} \\ \hline
          \textbf{Solar} & \textbf{Geomagnetic} & \textbf{Temporal} & \textbf{Solar} & \textbf{Geomagnetic} & \textbf{Temporal} \\ \hline 
          $F_{10}$,  $S_{10}$, & $a_p$,  \textit{Dst} & $t_1$,  $t_2$, & $F_{10}$,  $S_{10}$, & $a_{pA}$,  $a_p$,  $a_{p3}$, & $t_1$,  $t_2$, \\
          $M_{10}$,  $Y_{10}$, & & $t_3$,  $t_4$ & $M_{10}$,  $Y_{10}$, & $a_{p6}$,  $a_{p9}$,  $a_{p12-33}$, & $t_3$,  $t_4$ \\
          $F_{81c}$,  $S_{81c}$, & & & $F_{81c}$,  $S_{81c}$, &  $a_{p36-57}$,  \textit{Dst}$_A$, \textit{Dst}, & \\
          $M_{81c}$,  $Y_{81c}$ & & & $M_{81c}$,  $Y_{81c}$ & \textit{Dst}$_3$, \textit{Dst}$_6$,  \textit{Dst}$_9$, & \\
          & & & &  \textit{Dst}$_{12}$, \textit{Dst}$_{15}$, \textit{Dst}$_{18}$, \textit{Dst}$_{21}$ & \\ \hline
   \end{tabular}
\end{table}

In the JB\textsubscript{H} set, the geomagnetic indices are extensive in an effort to improve storm-time and post-storm modeling. The "A" subscript refers to the daily average, and the numerical subscripts refer to the value of the index that many hours prior to the epoch. The combination of two numbers references the number of previous hours the index is being averaged over (e.g. \textit{a\textsubscript{p12-33}} refers to the average \textit{a\textsubscript{p}} value between 12 and 33 hours prior to the prediction epoch). The authors found that using different time histories of \textit{a\textsubscript{p}} and \textit{Dst} (shown in Table \ref{t:inputs}) resulted in more optimal performance. For completeness, the results will also be shown using an input set that adopts the same time history for \textit{Dst} as the \textit{a\textsubscript{p}} time history in Table \ref{t:inputs}. This input set will be referred to as JB\textsubscript{H0}.

As the geomagnetic inputs in the JB\textsubscript{H} and JB\textsubscript{H0} sets are in a time series format, we tested the time-delay neural network (TDNN) approach. TDNNs use convolutional layers to absorb the information of the time history of a given input and make the model time invariant \citep{TDNN}. We ran tuners for the two historical input sets using the MSE loss function and did not see error reductions worthy of further investigation. Additionally, the architecture required for a TDNN adds significant complexity when implementing the NLPD and CRPS loss functions explained in Section \ref{sec:UQ}.

\subsection{Uncertainty Quantification}\label{sec:UQ}

Dropout is a regularization tool often used in machine learning (ML) to prevent the model from overfitting to the training data \citep{DO}. In standard feed-forward neural networks, each layer sends outputs from all nodes to those in the subsequent layer where they are introduced to weights and biases. Deep neural networks can have millions of parameters and thus are prone to overfitting. This causes undesired performance when interpolating or extrapolating.

Dropout layers use Bernoulli distributions, one for each node, with probability \textit{P}. This makes the model probabilistic since the distributions are sampled each time that a set of inputs are given to the model. If a sample is "true", the node's output is nullified, and the output of the layer is scaled based on the number of nullified outputs. Dropout is believed to make each node independently sufficient and not reliant on the outputs of other nodes in the layer \citep{DO2}.  

In traditional use, dropout layers are only activated during training to uphold the deterministic nature of the model. However, measures can be taken in order for this feature to remain activated during prediction making the model probabilistic. When passing the same input set to the model a significant number of times (e.g. 1,000), there is a distribution of model outputs for each unique input. This process is referred to as Monte Carlo (MC) dropout. Essentially, every time the model is presented with a set of inputs, random nodes are dropped out providing a different functional representation of the model. Gal and Ghahramani \citeyearpar{gal2016dropout} show that MC dropout is a Bayesian approximation of a Gaussian process.

In this case, the model uses the input set to predict the ten (10) PCA coefficients. Through the bootstrapping method, we estimate the mean and variance for each of the coefficients \citep{BS}. As the variance is different for each output/coefficient, this is a heteroskedastic problem. Each unique input can be passed to the model \textit{k} times and there will be \textit{k} $\times$ 10 outputs. The mean and variance are computed from those outputs with respect to the repeated axis, \textit{k}. The two loss functions used to improve uncertainty estimation (in addition to MSE) are NLPD and CRPS. NLPD is based on the logarithm of the probability density function (pdf) of the Gaussian distribution, and is shown in Equation \ref{eq4} \citep{NLPD,Camporeale_2021}.

\begin{equation} \label{eq4}
\begin{split}
NLPD(y,\mu,\sigma) = \frac{\left(y-\mu\right)^2}{2\sigma^2} + \frac{log(\sigma^2)}{2} + \frac{log(2\pi)}{2}
\end{split}
\end{equation}

In Equation \ref{eq4}, \textit{y} is the ground truth, $\mu$ is the mean prediction, and $\sigma$ is the predictive standard deviation, each being computed for all 10 outputs making it a heteroskedastic loss function. For clarity, the \textit{log} used in the NLPD loss function is the natural logarithm. The second loss function for uncertainty estimation is CRPS which is shown analytically in Equation \ref{eq5} \citep{CRPS}. The main difference between NLPD and CRPS is that CRPS is also includes the cumulative distribution function (cdf) of the Gaussian distribution as opposed to only the pdf,

\begin{equation} \label{eq5}
\begin{split}
CRPS(y,\mu,\sigma) = \sigma\left[\frac{y-\mu}{\sigma} \textrm{ erf}\left(\frac{y-\mu}{\sqrt{2}\sigma}\right) + \sqrt{\frac{2}{\pi}} \textrm{ exp}\left(-\frac{\left(y-\mu\right)^2}{2\sigma^2}\right) - \frac{1}{\sqrt{\pi}}\right].
\end{split}
\end{equation}

As with Equation \ref{eq4}, $\mu$ and $\sigma$ are computed using bootstrap methods, and this loss function is also heteroskedastic. An important aspect of using the loss functions described in Equations \ref{eq4} and \ref{eq5} is the preparation of the training data. The data is traditionally in the following form. The \textit{n} features are set up as the number of samples, with \textit{n\textsubscript{I}} denoting the number of inputs, resulting in the input shape dimension as (\textit{n} $\times$ \textit{n\textsubscript{I}}). The \textit{n} labels are the number of samples with   \textit{n\textsubscript{o}} the number of outputs, resulting in the output shape dimension as  (\textit{n} $\times$ \textit{n\textsubscript{o}}). To implement these loss functions, we stack each input and output by the number of MC samples, \textit{k}. This is a repeated axis, meaning all samples along this axis are identical. The resulting shapes of the features and labels are (\textit{n} $\times$ \textit{k} $\times$ \textit{n\textsubscript{I}}) and (\textit{n} $\times$ \textit{k} $\times$ \textit{n\textsubscript{o}}), respectively. This allows us to determine the mean and standard deviation for each unique input within the loss function.

\subsubsection{Latent Space UQ}

Since there are multiple models and loss functions to compare, we have to implement a metric to judge each model's ability to provide reliable uncertainty estimates. To do so, we modified a calibration error equation from Anderson et al. \citeyearpar{meaningful_UQ}, shown as 

\begin{equation} \label{eq6}
\begin{split}
\textrm{Calibration Error} = \sum^r_{i=1} \sum^m_{i=j} \Big|p(z_{i,j})-p(\hat{z}_{i,j})\Big|.
\end{split}
\end{equation}

In Equation \ref{eq6}, \textit{m} is the number of prediction intervals investigated, \textit{r} is the choice order of truncation for PCA (the number of model outputs), \textit{p(z)} is the expected cumulative probability, and \textit{p($\hat{z}$)} is the observed cumulative probability. The prediction intervals of interest in this work span from 5\% to 95\% with 5\% increments appended with 99\%. \textit{p($\hat{z}$)} is computed empirically shown in Equation \ref{eq7},

\begin{equation} \label{eq7}
\begin{split}
p\left(\hat{z}\right) = \frac{1}{n}\sum^n_{i=1} \mathbb{I}\left(\hat{z}_i^l<z_i<\hat{z}_i^u\right)
\end{split}
\end{equation}
where \textit{n} is the number of samples, $\mathbb{I}$ is the indicator function, $\hat{z}_i^l$, is the lower bound of the prediction interval, $\hat{z}_i^u$ is the upper bound of the prediction interval, and $z_i$ is the sample. To compute the bounds, we use the pair of equations given in Equation \ref{eq8},

\begin{equation} \label{eq8}
\begin{split}
\hat{z}_i^l = \mu - \mathbf{z}\sigma
\;\;\;\textrm{and}\;\;\;
\hat{z}_i^u = \mu + \mathbf{z}\sigma,
\end{split}
\end{equation}
where \textbf{z} is the critical value used for the prediction interval. This is calculated using the Gaussian CDF

\begin{equation} \label{eq9}
\begin{split}
\mathbf{z} = \sqrt{2}\textrm{ erf}^{-1}\left(PI\right)
\end{split}
\end{equation}
where \textit{PI} is the prediction interval of interest (e.g. 95\% or 0.95). 

\subsubsection{Density UQ}\label{sec:DUQ}

While latent space calibration is important because the model is trained on the PCA coefficients, determining the reliability of the model's predictions on the resulting density is the ultimate goal. To examine this, we look at the orbits of CHAMP and GRACE \citep{CHAMP, GRACE}. These satellites had onboard accelerometers from which mass density has been estimated \citep{CHden1,CHden2,sutton, doorn, EOF3, CHAGRA}. Licata et al. \citeyearpar{QualHASDM} recently compared the SET HASDM density database to both CHAMP and GRACE-A density estimates over the lifetime of their measurements.

We evaluate the model 1,000 times every 3 hours across the entire availability of CHAMP and GRACE measurements from Mehta et al \citeyearpar{CHAGRA} and interpolate them to the satellite locations. This allows us to compute the observed cumulative probability of the HASDM-ML model relative to the HASDM database. Due to the redundancy and computational expense, we only interpolate the model and database density every 500 samples (5,000 and 2,500 seconds for CHAMP and GRACE-A, respectively). This provides a wide view of the model's UQ capabilities. The CHAMP comparison uses 23,795 model prediction epochs (40.7\% of the total available data) with the density being interpolated to at least two satellite positions per prediction due to the cadence of this study. The GRACE comparison uses 24,602 model prediction epochs (42.1\% of the total available data) with the density being interpolated to at least four satellite positions per prediction. 

Geomagnetic storms are particularly difficult conditions to model accurately. Therefore, we look at four (4) geomagnetic storms from 2002-2004 where we can evaluate HASDM-ML's reliability across unique events. Two of the events take place in 2002, which is outside the training dataset, while the other two are from 2003 and 2004 and are seen in training. To conduct this assessment, we employ the methodology from the Licata et al. \citeyearpar{QualHASDM} storm-time case study. The model is evaluated over a six (6) day period encompassing a storm then interpolated to the CHAMP locations (10 second cadence). During each three-hour prediction period, the density grids remain constant. All 1,000 HASDM-ML density variations are then averaged across each orbit using the mean altitude to determine the orbital period. The mean and 95\% predition interval bounds are computed to compare to the corresponding HASDM densities and shown in the figure in an orbit-averaged form. In total, the six days amounts to 48 model prediction epochs which results in 51,840 interpolated densities (1,000 MC runs) from which we compute the observed cumulative probabilities.

\section{Results}\label{sec:results}

Upon running each input set with all three loss functions through individual tuners, the best ten models from each tuner are evaluated on the entire training, validation and test sets 1,000 times. The mean absolute error for each subset is computed based on the mean prediction transformed back to the density space. The calibration error score (Equation \ref{eq6}) is computed using the mean and standard deviation from the 1,000 runs in the latent space. The mean absolute error results from the best of the ten models for each input set/loss function are shown in Table \ref{t:Estat}.

\begin{table}[htb!]
	\fontsize{10}{10}\selectfont
    \caption{Mean absolute for the best model from each technique across training, validation, and test data.}
   \label{t:Estat}
        \centering 
   \begin{tabular}{c | c | c | c | c | c | c | c | c | c } 
      \hline 
          & \multicolumn{3}{c |}{\textbf{Training}} & \multicolumn{3}{c |}{\textbf{Validation}} & \multicolumn{3}{c}{\textbf{Test}} \\ \hline
          \textbf{Technique} & JB & JB\textsubscript{H} & JB\textsubscript{H0} & JB & JB\textsubscript{H} & JB\textsubscript{H0} & JB & JB\textsubscript{H} & JB\textsubscript{H0} \\ \hline
          \textbf{MSE} & 10.38\% & 8.73\% & 8.47\% & 12.00\% & 10.48\% & 9.91\% & 11.95\% & 10.71\% & 10.51\% \\ \hline
          \textbf{NLPD} & 10.07\% & 9.07\% & 8.81\% & 11.93\% & 10.69\% & 9.87\% & 11.41\% & 10.69\% & 10.05\% \\ \hline
          \textbf{CRPS} & 9.67\% & 8.64\% & 8.26\% & 11.56\% & 10.55\% & 9.69\% & 11.76\% & 10.43\% & 10.69\% \\ \hline
   \end{tabular}
\end{table}

The addition of historical geomagnetic indices clearly improves the model performance with error reductions upwards of 2\%. As mentioned in Section \ref{sec:regression}, the motivation for using the time series geomagnetic indices was to improve storm-time and post-storm performance. However, Table \ref{t:HASDM_samples} shows that these conditions account for a small subset of the data meaning the notable performance improvement with the JB\textsubscript{H} and JB\textsubscript{H0} input sets show that it likely improves the predictions across all conditions. In general, the CRPS models have the lowest error, and the JB\textsubscript{H0} models have the lowest error in terms of the input sets. Table \ref{t:Sstat} shows the calibration error score for the same models.

\begin{table}[htb!]
	\fontsize{10}{10}\selectfont
    \caption{Calibration error score for the best model from each technique across training, validation, and test data.}
   \label{t:Sstat}
        \centering 
   \begin{tabular}{c | c | c | c | c | c | c | c | c | c } 
      \hline 
          & \multicolumn{3}{c |}{\textbf{Training}} & \multicolumn{3}{c |}{\textbf{Validation}} & \multicolumn{3}{c}{\textbf{Test}} \\ \hline
          \textbf{Technique} & JB & JB\textsubscript{H} & JB\textsubscript{H0} & JB & JB\textsubscript{H} & JB\textsubscript{H0} & JB & JB\textsubscript{H} & JB\textsubscript{H0} \\ \hline
          \textbf{MSE} & 3.871 & 3.744 & 3.758 & 3.962 & 3.898 & 3.901 & 4.004 & 3.979 & 3.972 \\ \hline
          \textbf{NLPD} & 0.3404 & 0.3061 & 0.2794 & 0.3084 & 0.2506 & 0.2836 & 0.2212 & 0.1757 & 0.2790 \\ \hline
          \textbf{CRPS} & 0.3292 & 0.7933 & 0.4464 & 0.2270 & 0.4634 & 0.2735 & 0.2399 & 0.2387 & 0.2950 \\ \hline
   \end{tabular}
\end{table}

\begin{figure}[htb!]
	\centering
	\small
	\includegraphics[width=\textwidth]{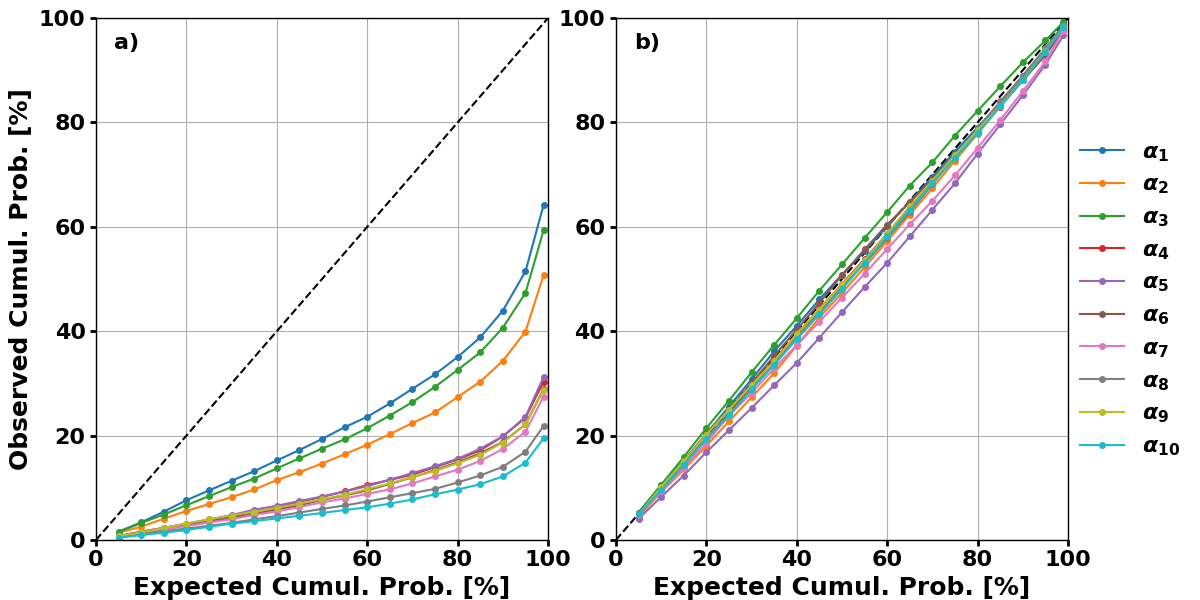}
	\caption{Expected vs observed cumulative probability (Pr.) of all 10 PCA coefficients for HASDM-ML on the test set using JB\textsubscript{H} + MSE (a) and JB\textsubscript{H} + NLPD (b).}
	\label{f:PCA_Cal}
\end{figure}

The incorporation of the custom loss functions reduce the calibration error score by an order of magnitude relative to models trained with MSE, which tend to underestimate the uncertainty. The best performing loss function, in regards to calibration, is NLPD. To choose the best overall model, we focus on the test performance as that data is completely independent from the training process. Furthermore, we value the most calibrated model as reliable uncertainty estimates are the most valuable asset for a thermospheric density model. The JB\textsubscript{H} + NLPD model is within 1\% of the error of all better-performing models, and it has the lowest test calibration error score with satisfactory values for the training and validation data. As the calibration error score is a composite of the scores from each coefficient, we show the calibration curves of all coefficients on the independent test set for the best JB\textsubscript{H} + NLPD model, in panel (b), alongside the best JB\textsubscript{H} + MSE model, in panel (a), for comparison in Figure \ref{f:PCA_Cal}.

The calibration curve in panel (b) for all PCA coefficients roughly follows the perfectly-calibrated 45$^{\circ}$ line with $\alpha_5$ being the only coefficient that prominently underestimates uncertainty. However, there is minimal contribution to the full-state (density) after the first few coefficients, so this should not greatly impact the resulting density. In sharp contrast, panel (a) shows the model trained with the MSE loss function is not nearly calibrated, as is evident in Table \ref{t:Sstat}. There is a strong underestimation of the uncertainty at all cumulative probability levels, because the model is not trained with any terms for its variance.

\begin{figure}[htb!]
	\centering
	\small
	\includegraphics[width=\textwidth]{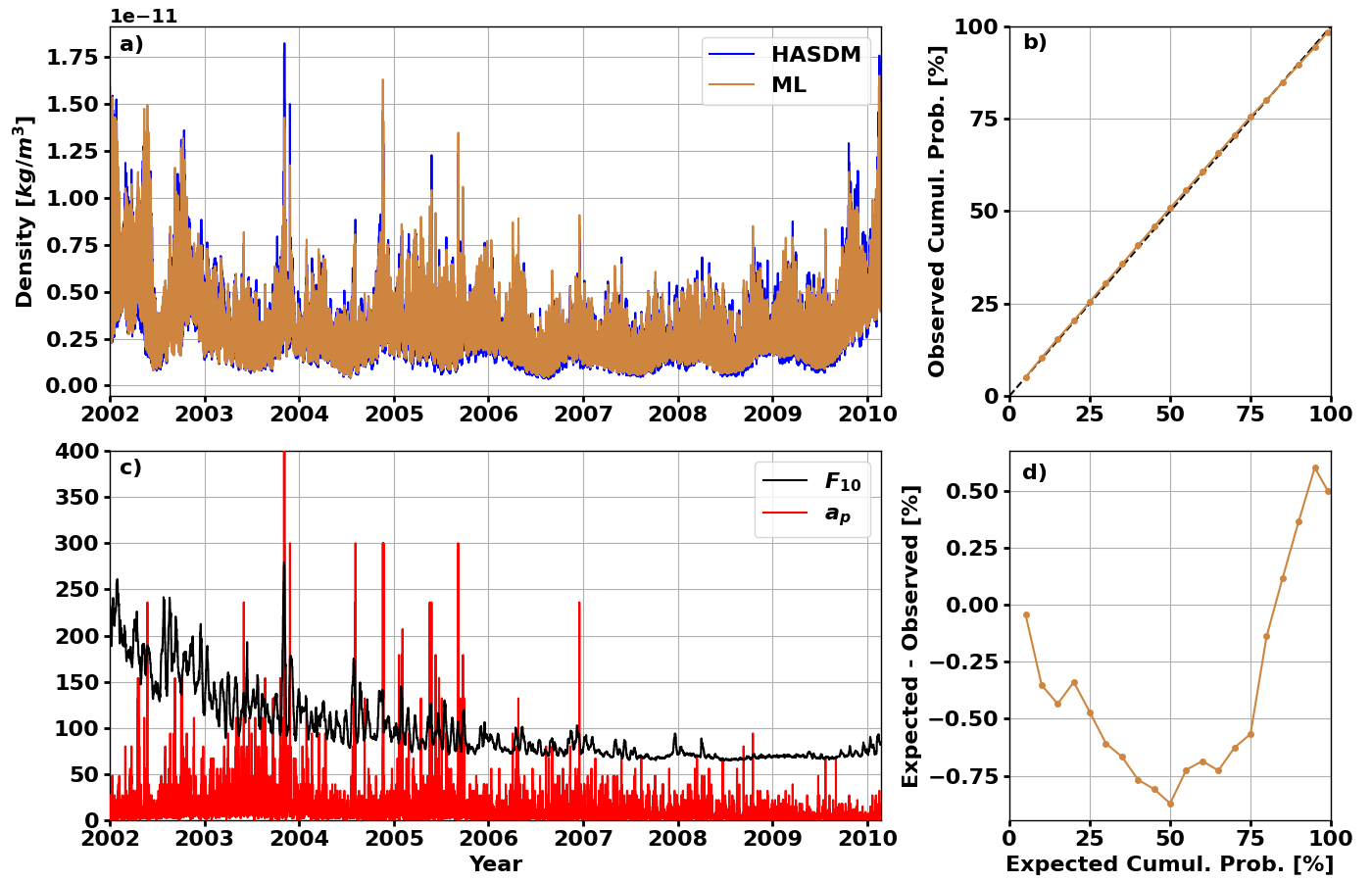}
	\caption{(a) shows the HASDM and mean HASDM-ML density interpolated to CHAMP positions, (b) shows the expected vs observed calibration curve, (c) shows \textit{F\textsubscript{10}} and \textit{a\textsubscript{p}} for the period corresponding to (a), and (d) shows the difference between expected and observed cumulative probability corresponding to (b). Discontinuities in (a) and (c) represent data gaps.}
	\label{f:CHAMP_Cal}
\end{figure}

The JB\textsubscript{H} + NLPD model shown in Figure \ref{f:PCA_Cal} will be the focus of all subsequent analyses and will be referred to as HASDM-ML. To investigate the model's reliability on density in an operational nature, we look at the orbits of CHAMP and GRACE-A each over eight year periods with a cumulative altitude range of 300 - 530 km. HASDM-ML was evaluated in three-hour increments from 2002-2011, and was interpolated to the satellite positions. The results for the CHAMP orbit are displayed in Figure \ref{f:CHAMP_Cal}.

\begin{figure}[htb!]
	\centering
	\small
	\includegraphics[width=\textwidth]{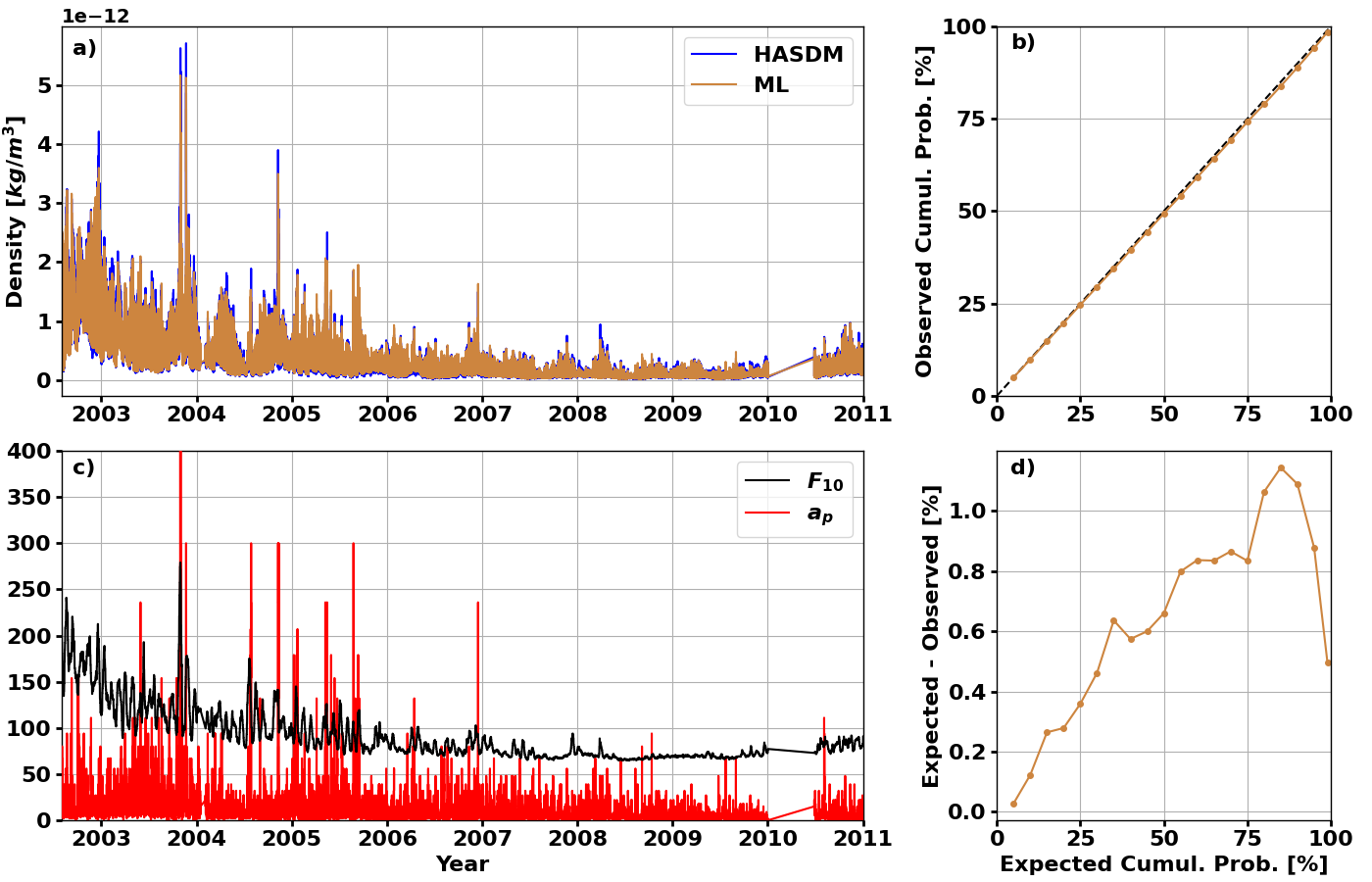}
	\caption{(a) shows the HASDM and mean HASDM-ML density interpolated to GRACE-A positions, (b) shows the expected vs observed calibration curve, (c) shows \textit{F\textsubscript{10}} and \textit{a\textsubscript{p}} for the period corresponding to (a), and (d) shows the difference between expected and observed cumulative probability corresponding to (b). Discontinuities in (a) and (c) represent data gaps.}
	\label{f:GRACE_Cal}
\end{figure}

Panel (a) shows a clear similarity in the HASDM and HASDM-ML mean densities. The ML model's density formulation does not appear to contain any major discrepancies. The surrogate ML model is imperfect in its mean prediction, as seen in Table \ref{t:Estat}, but panels (b) and (d) show that the density uncertainty is reliable. The calibration curve is exceptional with the observed cumulative probability being within 1\% of the expected value for all 20 cumulative probability levels tested. Figure \ref{f:GRACE_Cal} shows the same analysis along the GRACE-A orbit.

\begin{figure}[htb!]
	\centering
	\small
	\includegraphics[width=\textwidth]{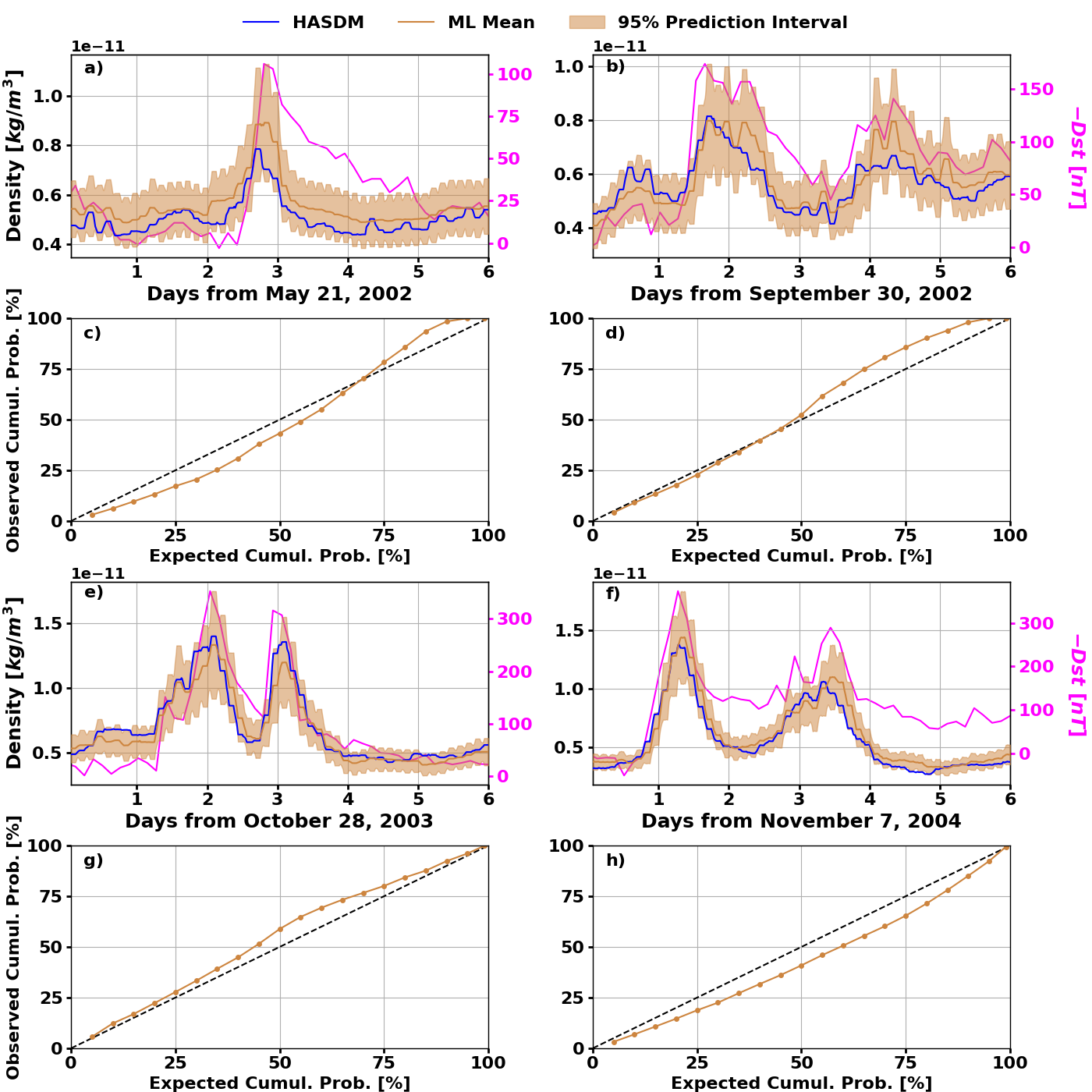}
	\caption{Panels (a), (b), (e), and (f) show HASDM and HASDM-ML mean orbit-averaged density for CHAMP's orbit across various geomagnetic storms. The shaded region represents the 95\% prediction interval for HASDM-ML, and -\textit{Dst} is shown on the right axis in each panel. Panels (c), (d), (g), and (h) show the calibration curves corresponding to panels (a), (b), (e), and (f), respectively.}
	\label{f:storms}
\end{figure}

Again, panel (a) shows that HASDM and HASDM-ML mean densities are quite similar. Panels (b) and (d) also demonstrate that although the densities are not identical, HASDM-ML provides uncertainty estimates that are reliable. Panel (d) reveals that at the higher GRACE altitudes, there is slightly less agreement with the expected and observed cumulative probabilities with the largest discrepancy being just over 1\%. To perform the simulations displayed in Figures \ref{f:CHAMP_Cal} and \ref{f:GRACE_Cal}, the model had to be evaluated 23,795,000 and 24,602,000 times for CHAMP and GRACE, respectively. These numbers come from the number of HASDM prediction epochs (Section \ref{sec:DUQ}) and the number of MC runs (1,000). HASDM-ML can perform these predictions in 17.27 seconds for CHAMP and 17.54 seconds for GRACE on a laptop with a NVIDIA GeForce GTX 1070 Mobile graphics card. Using CPU, the model takes 143 seconds for CHAMP and 152 seconds for GRACE. Figure \ref{f:storms} shows HASDM and HASDM-ML orbit-averaged densities during four geomagnetic storms with prediction intervals and the associated calibration curves.

Across all of the storms investigated, the mean prediction of HASDM-ML follows the trend of HASDM density. Even when the model deviates, HASDM densities are mostly captured by the uncertainty bounds. Panels (a) and (b) represent storms not contained in the training dataset which show that HASDM-ML is well-generalized, even during these highly nonlinear events. The calibration curves corresponding to each event show the robust nature of HASDM-ML's uncertainty estimates. While the observed cumulative probability values deviated from the expected values, these are highly nonlinear periods where density models tend to be unreliable.

\subsection{HASDM-ML Performance Metrics}\label{sec:error_stats}

To assess the conditions in which HASDM-ML can improve, the global mean absolute errors are computed as a function of space weather conditions. The results are shown in Table \ref{t:model_stats} and the number of samples in each bin can be found in Table \ref{t:HASDM_samples}.

\begin{table}[htb!]
	\fontsize{10}{10}\selectfont
    \caption{Mean absolute error across global grid for HASDM-ML as a function of space weather conditions.}
   \label{t:model_stats}
        \centering 
   \begin{tabular}{c | c | c | c | c | c} 
      \hline 
          & $\mathbf{F_{10} \leq 75}$ & $\mathbf{75 < F_{10} \leq 150}$ & $\mathbf{150 < F_{10} \leq 190}$ & $\mathbf{F_{10} > 190}$ & \textbf{All} $\mathbf{F_{10}}$\\ \hline
         $\mathbf{a_p \leq 10}$ & 8.96\% & 9.78\% & 9.97\% & 9.14\% & 9.50\% \\ \hline
         $\mathbf{10 < a_p \leq 50}$ & 9.76\% & 10.05\%  & 10.87\%  & 9.90\% & 10.09\% \\ \hline
         $\mathbf{a_p > 50}$ & 15.35\% & 12.86\% & 13.23\%  & 12.55\% & 13.01\% \\ \hline
         \textbf{All} $\mathbf{a_p}$ & 9.12\% & 9.92\% & 10.36\% & 9.55\% & 9.71\% \\ \hline
   \end{tabular}
\end{table}

The results from Table \ref{t:model_stats} show that HASDM-ML is robust to different \textit{F\textsubscript{10}} and \textit{a\textsubscript{p}} ranges when \textit{a\textsubscript{p}} $\leq$ 50.  The only conditions in which the mean absolute error exceeds 11\% is when \textit{a\textsubscript{p}} > 50, which only accounts for 1.80\% of the samples. This shows that more samples are required for this specific condition in both the training and evaluation phases. The last row contains the errors only as a function of \textit{F\textsubscript{10}} which shows that across all four solar activity levels, the error deviates by only 1.24\%. The bottom-right cell shows that the error across all 20 years of available data is only 9.71\%.

\section{Summary}\label{sec:con}

In this work, we developed a surrogate ML model for the SET HASDM density database with robust and reliable uncertainty estimation capabilities. Principal component analysis was selected for dimensionality reduction due to its versatile nature. A Bayesian search algorithm was leveraged to identify an optimal architecture for each input set and loss function tested. We found that of the nine input-loss function combinations explored, the combination of a JB2008 input set with historical geomagnetic drivers and the heteroskedastic NLPD loss function resulted in the most comprehensive model. This model, HASDM-ML, has 9.07\% error across the 12 year training set and an average 10.69\% error over the combined 8 year validation/test sets. It also had the lowest calibration error score on the test set, meaning it was the most calibrated to independent data. We also compared its calibration curves for each output across the test set to that of the MSE model with the same inputs. This showed that the MSE model considerably underestimated the uncertainty while the NLPD model was well-calibrated across the 10 outputs. 

Upon selecting HASDM-ML, we evaluated its uncertainty capabilities across the entire orbits of CHAMP and GRACE-A, both using almost half of the time span of the dataset. This assessment showed that the mean prediction at the satellite locations closely matched that of the HASDM dataset. Across all 20 prediction intervals tested over this period, the model provided an observed cumulative probability that never deviated more than 1\% of the expected value for CHAMP's orbit and never deviated more than 1.15\% for GRACE-A's orbit. A separate storm-time evaluation unveiled that across four storms, HASDM-ML provides similar density to HASDM and its uncertainty estimates remained robust and reliable. The results from the density calibration tests are significant, because no thermospheric density model to date provides uncertainty estimates. Additionally, uncertainty estimates themselves are not meaningful unless they are well-calibrated, and HASDM-ML is able to provide that.

\section{Limitations and Future Work}

To further improve HASDM-ML, additional data can be introduced, particularly in the coming solar maximum. As seen in Section \ref{sec:error_stats}, periods of \textit{a\textsubscript{p}} > 50 were the source of the highest overall error and the introduction of additional storms could reduce that error. Hierarchical modeling is another approach to potentially combat this limitation. A nonlinear dimensionality reduction method could also improve performance, especially in modeling the highly nonlinear storm response. The developed model can be used in research (e.g. to study historical storms where HASDM outputs are unavailable) as well as in an operational setting. Another area of future work could be to develop a nonlinear dynamic ROM based on the SET HASDM density database. While the background model is static, the assimilative framework represents the dynamic thermosphere, and a dynamic ROM could better model certain phenomena and provide a framework for further data assimilation \citep{DAframework}.

\section{Data Statement}

Requests can be submitted for access to the SET HASDM density database at \url{https://spacewx.com/hasdm/} and all reasonable requests for scientific research will be accepted as explained in the rules of road document on the website. The historical space weather indices used in this study can also be found at \url{https://spacewx.com/jb2008/}. Original CHAMP and GRACE density estimates from Mehta et al. (2017) can be found at \url{http://tinyurl.com/densitysets}.

\section{Acknowledgment}

This work was made possible by NASA Established Program to Stimulate Competitive Research, Grant $\#$80NSSC19M0054. SET and WVU gratefully acknowledge support from the NASA SBIR contract $\#$80NSSC20C0292 for Machine learning Enabled Thermosphere Advanced by HASDM (META-HASDM). The authors would like to thank the anonymous reviewers for all of their time and effort. 

\bibstyle{AAS_Publications}   
\bibliography{references}   

\end{document}